# Detecting Anomalous Process Behaviour using Second Generation Artificial Immune Systems

Jamie Twycross[1], Uwe Aickelin[2] and Amanda Whitbrook[2]

[1] *ASAP Research Group, School of Computer Science, University of Nottingham, UK*
[2] *IMA Research Group, School of Computer Science, University of Nottingham, UK*
*E-mail: jpt@cs.nott.ac.uk; uxa@cs.nott.ac.uk; amw@cs.nott.ac.uk*

Artificial Immune Systems have been successfully applied to a number of problem domains including fault tolerance and data mining, but have been shown to scale poorly when applied to computer intrusion detection despite the fact that the biological immune system is a very effective anomaly detector. This may be because AIS algorithms have previously been based on the adaptive immune system and biologically-naive models. This paper focuses on describing and testing a more complex and biologically-authentic AIS model, inspired by the interactions between the innate and adaptive immune systems. Its performance on a realistic process anomaly detection problem is shown to be better than standard AIS methods (negative-selection), policy-based anomaly detection methods (systrace), and an alternative innate AIS approach (the DCA). In addition, it is shown that runtime information can be used in combination with system call information to enhance detection capability.

## 1 INTRODUCTION

This paper is concerned with the classification performance of a novel Artificial Immune System (AIS) on a process anomaly detection problem. The novel AIS (the `tlr` algorithm) incorporates mechanisms inspired by both the innate and adaptive biological immune systems, and produces a very low false positive rate when detecting attacks on an FTP server. As with many other process anomaly detection systems, system call information is used as one source of input data. However, another novel aspect of the `tlr` algorithm is the use of runtime statistics (such as process memory and file usage) as context signals that form additional sources of input data. This aspect builds on the idea of gray-boxing, a term introduced several years ago [8] to denote intrusion detection systems that use runtime information as well as system call information.

The rest of the paper is structured as follows. Section 2 provides some essential background information on intrusion and process anomaly detection, reviewing and describing the various approaches. It also introduces some fundamental AIS concepts including the notion of first and second generation AIS algorithms. A detailed explanation of the `tlr` algorithm's architecture is given in Section 3, and Section 4 describes how a test dataset (`wuftpd`) is created and how the normal and anomalous test data is constructed, i.e., how the input system calls and context signals used by the `tlr` algorithm are gathered in practice. Section 5 reports on the experimental procedures adopted and the results are presented and discussed in Section 6; in particular, the `tlr` algorithm's performance is compared with those of several other classifiers and anomaly detection approaches. Section 7 concludes the paper.

## 2  BACKGROUND

### 2.1  Process Anomaly Detection

A process is a running instance of a program, and on modern multitasking operating systems many processes are effectively running simultaneously. For example, a server may be running a web server, email servers and a number of other services. A single program executable, when run, may create several child processes by forking or threading, and is then known as the parent process of those child processes; web servers typically start child processes to handle individual connections once they have been received. Child processes themselves may create children, sometimes generating a complex process tree derived from a single parent-process node, created when the executable is first run. The operating system is responsible for managing the execution of running processes, and associates a number with each one. This is called the process identifier (PID), as it uniquely identifies each process. When a process is started, the operating system associates other metadata with it too, such as the user who started it, and the PID of the parent process that created it. The operating system also allocates resources to running processes, including memory (which stores the executable code and data) and file descriptors, which identify files or network sockets that belong to the process.

A number of host-based Intrusion Detection Systems (IDSs) have been built around monitoring running processes to detect intrusions. In general, these IDSs collect information about a running process from a variety of sources, including from log files created by the process, or from other information gathered from the operating system. The general idea is that by observing what the process is currently doing, for example by looking at its log files, it is possible to tell whether the process is behaving normally or has been subverted by an attack. Whilst log files are an obvious starting point for such systems, and are still an important component in a holistic security approach, it is fairly easy to execute attacks which do not cause any logging to take place, and so evade detection. Because of this, much research effort has been directed towards the use of other data sources, usually collected by the operating system. Of these, system calls (syscalls) have been the most favoured approach.

A syscall is a low-level mechanism by which an application requests system services such as peripheral I/O or memory allocation from an operating system. As a process runs it cannot usually directly access memory or hardware devices; instead, the operating system manages these resources and provides a set of functions, called syscalls, which processes can call to access these services. On modern Linux systems there are around 300 syscalls, accessed via wrapper functions in the libc library.

### 2.2 Process Anomaly Detection Systems

Due to space constraints, this section focuses on syscall-based IDSs. The `systrace` system of Provos [18] is a syscall-based confinement and IDS for Linux, BSD and OSX systems. The IDS works by using a kernel patch that inserts various hooks into the kernel to intercept syscalls from the monitored process, and the user has to specify a syscall policy, i.e. a whitelist of permitted syscalls and arguments. The system can be run either automatically to deny and log all syscall attempts not permitted by the policy, or to prompt a user to permit or deny the syscall graphically. The latter mode can also be used to add syscalls to the policy, adjusting it before using it in automatic mode. Initial policies for a process are obtained by using templates or by running `systrace` in automatic policy-generation mode, where the monitored process is run under normal usage conditions, and permit entries are created in the policy file for all the syscalls made by the process. The policy specification also allows some matching of syscall arguments as well as syscall numbers. The system's automatic policy-generation approach is used as a baseline comparison for the `tlr` algorithm presented in this paper, which can be seen as a more sophisticated and dynamic alternative to `systrace`.

In [8], Gao et al. introduce a new model of syscall behaviour called an execution graph. An execution graph is a model that accepts approximately the same syscall sequences as a model built on a control flow graph. However, the execution graph is constructed from syscalls gathered during normal execution, as opposed to a control flow graph, which is derived from static analysis. In addition to system call number, stack return addresses are also gathered and used in construction of the execution graph. The authors also introduce a course-grain classification of syscall-based IDSs into white-box, black-box and gray-box approaches. Black-box systems build their models from a sample of normal execution using only system call number and argument information. Gray-box approaches, as with black boxes, build their models from a sample of normal execution but, as well as using syscall information, also use additional runtime information. White-box approaches do not use samples of normal execution, but instead use static analysis techniques to derive their models. The `tlr` algorithm, in Gao's terms, is a gray-box approach, and is complementary to the specific gray-box approach described by Gao, exploring different sources of runtime information other than stack return addresses. Specifically, `tlr` uses the memory and file-usage levels of the executing application.

Forrest, Hofmeyr, Somayaji and other researchers at the University of New Mexico have developed several immune-inspired learning-based approaches. In [14], Forrest et al. evaluate a realtime system that detects anomalous processes by analysing sequences of system calls. Syscalls generated by an application are grouped together into sequences, in this case sequences of six consecutive syscalls. A database of normal sequences is constructed and stored as a tree during training. Sequences of syscalls are then compared to this database using a Hamming distance metric, and a sufficient number of mismatches generates an alert. Somayaji [21] uses a similar approach to develop the immune-inspired pH intrusion prevention system, which detects and actively responds to changes in program behaviour in realtime. If an anomaly is detected, execution of the process that produced the syscalls is delayed for a period of time. The work presented in this paper differs from these approaches in that the `tlr` algorithm does not actively respond to misbehaving processes as in [21], it only generates alerts. Also, `tlr` bases its alerts on the simple syscall number combined with other runtime information (to improve detection capability), as opposed to the more complex representations of syscalls used by Forrest et al.

### 2.3 Artificial Immune Systems

The field of Artificial Immune Systems (AIS) began in the early 1990s with a number of independent groups conducting research that used the biological immune system as inspiration for solutions to problems in other domains. AISs have been built for a wide range of applications including document classification, fraud detection, and network- and host-based intrusion detection [6]. Specifically of relevance to the work here are AIS approaches to intrusion detection, which are reviewed by Aickelin et al. [5]. For example, a negative selection AIS algorithm is used in the process anomaly detection systems built by Forrest et al. [7]. AISs have met with some success and in some cases have rivalled or bettered existing statistical and machine learning techniques [13].

Immunology textbooks generally characterise the innate and adaptive immune systems as separate. The adaptive system is described as capable of specific recognition and remembrance of antigen, while the innate system is seen mainly as a first line of defence and rapid-response mechanism. Understandably perhaps, from this perspective, a computer scientist might view the adaptive immune system as having more interesting properties such as learning and memory. However, this view of the immune system as two discrete systems does not reflect the intensive research and reassessment of the role of the innate immune system conducted over the last decade, as evidenced by the large number of papers published in immunology journals [9]. This research has uncovered many mechanisms by which the innate immune system interacts with the adaptive immune system, and has highlighted the role of the innate immune system as controller of the adaptive system. In other words, the protection afforded to the host by the immune system *as a whole* arises from mechanisms of the innate *and* adaptive immune systems, which form an *integrated system*. This new understanding of the structure and control of the immune system has led computer scientists to rethink the way in which they design their AISs; i.e. the importance of the innate immune system in AISs should mirror its worth in the biological organism.

One of the aims of this work is to show the value of considering the biological immune system as composed of interacting innate and adaptive subsystems when attempting to design a realistic and profitable AIS model. Over the last few years a number of design principles have been developed for constructing what are termed second generation AISs [25]. These employ algorithms inspired by both the biological innate and adaptive immune systems, as opposed to first generation AISs, which employ algorithms inspired only by the adaptive immune system. A software system called `libtissue` has also been developed. This allows researchers to implement second generation AISs as multiagent systems and to analyse their behaviour when they are applied to real-world problems [23, 26]. With the work here and other work, the aim is to show how second generation AISs can overcome some of the problems that have been attributed to first generation AISs, for example accuracy and scalability [26].

## 3 THE `tlr` ALGORITHM

The `tlr` algorithm is inspired by current immunological understanding of the interactions between two classes of immune cell: dendritic cells (DCs) and T cells (TCs). In particular, `tlr` uses a model of DC polarisation of T helper cells, based on the work of Kapsenberg [16]. Section 3.1 describes the biological theory that relates to these cells, their interactions and their environments, and Section 3.2 shows how these ideas have been abstracted to form the `tlr` algorithm - a working model of the innate and adaptive immune subsystems. Section 3.3 explains how the model has been used to create a host-based grey-box IDS.

### 3.1 The Underlying Biology

The adaptive immune system possesses two major types of lymphocytes that detect and respond to antigens, B cells (BCs) and T cells (TCs). TCs, the focus of this section, are responsible for the cell-mediated immune response and possess receptors that can be thought of as complex sensors specific to features of antigens. For adaptive immune cells these receptors are somatically generated (created by a complex process of gene segment rearrangement within the cell) and are termed *variable-region* receptors, as each is specific for a particular protein sequence. Variable-region receptors are selected for over the lifetime of the organism by processes such as clonal expansion, deletion or anergy and are under *adaptive* not evolutionary pressure. TCs recognise a non-self target only after antigens have been processed and presented in combination with a self receptor called a major histocompatibility complex (MHC) molecule.

TCs begin life in a naive state, in the lymph node, and there are two major subtypes; the killer TC and the helper TC. Killer TCs only recognize antigens coupled to Class I MHC (MHCI) molecules and are specialized in attacking cells of the body infected by viruses and sometimes by bacteria. They can also attack cancer cells. In contrast, helper TCs only recognize antigens coupled to Class II (MHC2) molecules and are responsible for regulation of both the innate and adaptive immune responses. They help determine which type of immune response the body will make to a particular antigen. Most antigens are T-dependent, meaning that two signals are necessary before the cell is attacked. The first signal comes from cross linking of the BC receptor and antigen and the second signal comes from co-stimulation provided by the helper TC. Co-stimualtion occurs when antigen presenting cells (APCs), for example DCs, present antigen on their MHC2 molecules. When these are recognized by helper TCs, the helper TC is activated and releases cytokines and other stimulatory signals that cause the activity of macrophages, killer TCs and BCs.

In contrast to the adaptive case, the receptors of innate system cells are entirely *germline-encoded*, in other words their structure is determined by the genome of the cell and has a fixed, genetically-determined specificity. Unlike adaptive system cells, they recognise a set of ligands under *evolutionary* pressure. One key group of innate receptors is the *pattern recognition receptor* (PRR) superfamily which recognises evolutionary-conserved *pathogen-associated molecular patterns* (PAMPs). PRRs do not recognise a specific feature of a specific pathogen as variable-region receptors do, but instead recognise common features or products of an entire class of pathogens. Thus, innate immune system receptors are termed *non-specific*, while adaptive immune system receptors are termed *specific*. The toll-like receptor (TLR) family of PRRs is the best characterised.

Recently a lot of research effort has been directed towards understanding how the innate immune system mediates the quality of an adaptive immune system response [9, 15]. Simplistically, this is concerned with understanding how the DCs interact with the TCs to prevent them from becoming active in the presence of self-antigen. DCs are generated in the bone marrow and initially reside as immature cells in the epithelia of the skin and mucosal tissue.

Their main functions are phagocytosis (the capture of complex molecules and entire cells from their surrounding environment) and antigen presentation. They collect the antigen through antigen receptors (AgRs) and express a special set of TLRs, which respond to PAMPs and are also activated by host-derived endogenous molecules (danger signals) that are produced when tissue is damaged. Periodically, DCs migrate to the draining lymphoid tissues where they halt phagocytosis, display the peptides they have collected, and interact with naive TCs. When their TLRs have become activated and antigen have been detected they differentiate into mature DCs and immediately migrate. They also migrate when they have reached their maximum lifespan and have detected antigen, in which case they differentiate into semimature DCs. Both semimature and mature DCs express antigen producers (AgPs), which give other cells access to the antigen they have collected. Mature DCs alone produce IL-12, which is used by naive TCs to differentiate between mature and semimature DCs. Essentially, the presence of biological danger signals causes a DC to present its antigen in a mature, immunogenic context, causing TC activation, whereas their absence causes the antigen to be presented in a semimature, tolerogenic context, and the TC is deleted.

In this way, DCs match the quality of the adaptive immune effector response to the nature of the antigen. They are therefore vital in the control of the adaptive immune system and the generation of tolerance to antigen in peripheral tissue. Through the production of a range of cytokines, DCs control the activation and proliferation of TCs and BCs, and determine the qualitative and quantitative nature of the adaptive immune response.

### 3.2 The `tlr` Model

As in Kapsenberg's model [16], DCs can either be immature, semimature or mature, and TCs either naive or activated. Furthermore, in the `tlr` model, cells exist within either the extralymphoid tissue compartment or lymph node compartment, and cell types are restricted to particular compartments. This is shown schematically in Figure 1. Immature DCs and activated TCs are only found in the extralymphoid compartment, and semimature DCs, mature DCs and naive TCs in the lymph node compartment.

Just as in the biological system, the model cells are not immortal but live for a certain period of time, which leads to a fluctuation in population levels for certain cell types. The number of immature DCs and naive TCs are purposefully maintained at constant levels, i.e. whenever an immature DC matures, it is immediately replaced by another immature DC, and whenever a naive TC dies it is immediately replaced by another naive TC. However, the population levels of semimature DCs, mature DCs and activated TCs are not fixed and are homeostatically determined by the cells of `tlr` itself and their environment. In order to detect anomalies the level of activated TCs is monitored, and if any activated TCs are produced an alert is generated. Homeostatic determination of cell numbers is found to be particularly useful since, during periods of normal usage, cell numbers are generally kept at low levels, which reduces the computational cost of the algorithm.

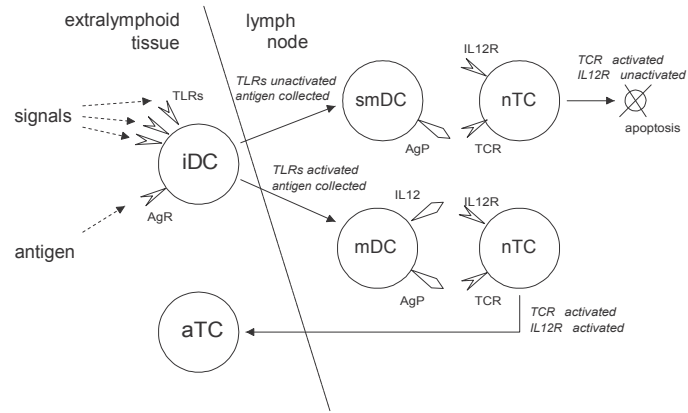

FIGURE 1
A schematic representation of the `tlr` algorithm. For DCs i = immature, sm = semimature and m = mature. For TCs a = activated, n = naive.

DCs begin life in the extralymphoid compartment, in an immature state, where they collect antigen (syscalls) through their AgRs and observe the levels of context signals (other system information) through their TLRs. If their TLRs become activated by an appropriate context signal and at least one antigen has been collected, immature DCs differentiate into mature DCs and traffic to the lymph node compartment. However, if an immature DC reaches its maximum lifespan without its TLRs being activated but has collected at least one antigen, it differentiates into a semimature DC and traffics to the lymph node compartment. Immature DCs which reach the end of their lifespan without collecting antigen remain in the extralymphoid tissue. Here, activating levels of context signals are analogous to biological danger signals, which cause a DC to present its antigen in the mature, immunogenic context.

TCs begin life as naive TCs in the lymph node compartment and bind with a number of semimature or mature DCs per cycle, provided there is at least one semimature or mature DC in this compartment. The DC is chosen randomly and uniformly from the complete population of semimature and mature DCs, so if the number of mature DCs is greater than the number of semimature DCs, a naive TC will have a greater probability of binding with a mature DC than a semimature DC and vice versa. If a naive TC successfully binds with a DC then it will examine the antigen the DC has collected as an immature DC in the extralymphoid compartment (and is displaying on its AgPs). If a match occurs and the bound DC is a semimature DC then the naive TC will be deleted and the semimature DC will receive a stay-alive signal. This stay-alive signal resets the number of iterations the semimature DC has existed for to zero. If a match occurs and the bound DC is a mature DC then the naive TC will be activated and will clone a single activated TC. The activated TC will then traffic back to the extralymphoid compartment and the naive TC will remain in the lymph node compartment. In this case, both the naive TC and mature DC receive stay-alive signals, resetting the number of iterations these cells have existed for to zero, based on biological mechanisms of reverse signalling. Detailed pseudocode for the `tlr` algorithm is provided in Section 3.4.

### 3.3 The Use of `tlr` as an IDS

The TLRs on DCs are activated by certain context signal values analagous to biological danger signals. Here, system information (other than syscall number) is used, (discussed shortly in Section 4). In contrast, AgRs on naive TCs are activated by certain antigen values, which are the syscall numbers. In order to determine which signal and antigen values activate these receptors `tlr` needs to be provided with a set of training data consisting of a sample of normal instances only. Once this is provided, all of the normal antigen and danger signal values are extracted and stored, and then a new set of permissible AgR values is created by removing all the antigen observed in the training set from the set of all possible antigen values (around 350 in the case of syscall numbers). In a similar way, TLRs are only activated by signal levels not seen in the training set. However, since context signals are represented as real numbers, the set of all possible values for a particular context signal need not be finite. Additionally, TLRs, unlike AgRs, are not specific for one particular value, but rather any value not seen in the training set. The signals used here (see Section 4) are all integers from finite sets, so this scheme works well. However, the levels of other real-valued signals would need to be discretised for this scheme to work effectively with them.

After training has taken place and the algorithm is tested, AgRs are generated by choosing a value at random from the permissible set of antigen, and assigning it as the receptor's lock. Exact matching is used, so that the AgR is only activated if its lock matches that of an antigen presented on an AgP of a DC. This is equivalent to the naive TC undergoing negative selection on the antigen in the training set, and produces naive TCs with AgRs that will never match an antigen seen in the training set.

Syscalls are collected and grouped temporally, i.e. those which occur around the same time period are collected and stored, and the external signals sensed by immature DCs also group the syscalls contextually; semimature DCs associate a normal context to the syscalls they have collected, while mature DCs associate an attack context with these syscalls. The addition of contextual grouping as well as temporal grouping has two effects on the repertoire of AgRs. First, non-deterministic interactions between semimature DCs and naive TCs allows `tlr` to continue to censor its AgR repertoire after the training phase, i.e. AgRs that are specific for syscalls produced during normal sessions (but not present during training) are quickly removed.

This peripheral tolerance mechanism helps reduce the number of false positives. Second, non-deterministic interactions between mature DCs and naive TCs promote the survival of naive TCs that successfully match antigen presented by mature DCs. These naive TCs are given a stay-alive signal by the mature DCs and remain within the population of naive TCs for longer. Since the population size of naive TCs is constant, the rate of entry of new naive TCs into the current naive TC population is reduced. Therefore, the contextual information provided by mature DCs serves to help maintain a population of successful naive TCs, and reduce the number of incoming naive TCs. The `tlr` algorithm is formalized in the pseudocode given below, and the `libtissue` parameters used are given in Table 4 in Appendix A. A full justification of the selection of parameters is provided in [24], Chapter 8.

### 3.4 `tlr` Pseudocode

Pseudocode for the `tlr` algorithm, see Table 4 in Appendix A for parameter values.

```
SUBROUTINE immature dc cell cycle callback
    IF cell iterations >= cell_lifespan_1 THEN
        IF cell has collected any antigen THEN
            SET cell type to semimature dc
            SET cell cycle to semimature dc cell cycle
            SET cell iterations to 0
            SET number of antigen producers to num_antigen_producers_1
            SET number of antigen receptors to 0
            SET number of cytokine receptors to 0
            add a new immature dc to tissue compartment
            RETURN
        ELSE
            replace cell with a new immature dc
            RETURN
        ENDIF
    ENDIF
    FOR all cytokine receptors DO # num_cytokine_receptors_1
        IF cytokine receptor activated THEN
            IF cell has collected any antigen
                THEN SET cell type to mature dc
                SET cell cycle to mature dc cell cycle
                SET cell iterations to 0
                SET number of antigen producers to num_antigen_producers_1
                SET number of antigen receptors to 0
                SET number of cytokine receptors to 0
                add a new immature dc to tissue compartment
                RETURN
            ENDIF
        ENDIF
    ENDFOR
ENDSUBROUTINE

SUBROUTINE naive tc cell cycle callback
    IF cell iterations >= cell_lifespan_2 THEN
        replace cell with a new naive tc
        RETURN
```

```
        ENDIF
        FOR all vr receptors DO # num_vr_receptors_2
            IF vr receptor activated THEN
                IF vr receptor activated by antigen on semimature dc
                    THEN SET semimature dc iterations to 1
                    replace cell with a new naive tc
                    RETURN
                ENDIF
                IF vr receptor activated by antigen on mature dc THEN
                    SET mature dc iterations to 1
                    SET cell type to activated tc
                    SET cell cycle to activated tc cell cycle
                    SET cell iterations to 0
                    SET number of cell receptors to 0
                    SET number of vr receptors to 0
                    add a new naive tc to tissue compartment
                    WRITE matched antigen to log file
                    RETURN
                ENDIF
            ENDIF
        ENDFOR
ENDSUBROUTINE

SUBROUTINE semimature dc cell cycle callback
    IF cell iterations >= cell_lifespan_3 THEN
        remove semimature dc from tissue compartment
    ENDIF
ENDSUBROUTINE

SUBROUTINE mature dc cell cycle callback
    IF cell iterations >= cell_lifespan_4 THEN
        remove mature dc from tissue compartment
    ENDIF
ENDSUBROUTINE

SUBROUTINE activated tc cell cycle callback
    IF cell iterations >= cell_lifespan_5 THEN
        remove activated tc from tissue compartment
    ENDIF
    FOR each antigen in tissue compartment
        DO IF vr receptor matches antigen
        THEN
            SET cell iterations to 0
            BREAKFOR
        ENDIF
    ENDFOR
ENDSUBROUTINE
```

Greensmith [10–12] has also used `libtissue` to implement an immune-inspired process anomaly detection system. This algorithm, called the DCA, is inspired by biological DCs and is similar to the `tlr` algorithm in its use of `libtissue` and models of biological DC activation and maturation. However, there are several important differences between `tlr` and the DCA. Although both systems incorporate the use of signals to govern DC behaviour, they do so in different ways. For example, the DCA might identify CPU usage as belonging to the class of safe signals and memory usage to the class of danger signals. Due to the way DCs integrate and process input signals, the class an input signal belongs to has a differential effect on DC maturation. If the classes of CPU and memory usage signals are interchanged, the algorithm behaves differently. In contrast, the `tlr` algorithm does not differentiate between different classes; all input signals have the same effect on DC maturation, i.e. they cause immature DCs to become mature DCs if the signal level is not observed during training. In this sense, the DCA uses a finer-grain model of input signals than `tlr`.

Another important difference relates to the way in which anomalies are detected by the two systems. The DCA observes DCs to determine if a process is behaving anomalously, whereas `tlr` uses DCs to control TCs and observes TCs to determine anomalous process behaviour. While the DCA concentrates on developing a system based entirely on DCs, `tlr` focuses on combining a DC-based algorithm with current adaptive AIS algorithms. This is a result of different motivations for developing the DCA and the `tlr` algorithm. The DCA was developed to explore better methods of modelling DCs in AISs, whereas `tlr` was developed to explore the construction of second generation AISs that incorporate both innate and adaptive immune system mechanisms. Lastly, an essential part of `tlr` is the training phase in which normal usage is used to establish activating levels of external signals, as well as permissible values for AgRs. The DCA does not use a training phase, but instead uses heuristics derived from observations of the biological immune system and the computer system being protected to determine activating levels of signals. The DCA has been tested on the same dataset used here and its results [10] are presented for comparison with the `tlr` algorithm in Section 6.

## 4  DATASETS

### 4.1  Network Architecture

In order to derive a suitable test dataset, a small experimental network with two hosts is set up. One host, the target, runs software, in this case a Redhat 6.2 server with a number of vulnerabilities, and `wuftpd` is started at boot time. The other host acts as a client which interacts with the target machine, either attempting to exploit its vulnerabilities or simulating normal usage. In order to gather the actual data, i.e. process syscall information and context signals, the target system is instrumented: the FTP server executable is wrapped with `strace` [18], which logs all the syscalls made by `wuftpd` and its children. At the same time, a `process_monitor` is started, which monitors a process and all of its child processes at regular intervals. In order to see a useful resolution in the signals, a monitoring interval of one tenth of a second is used. This method is found to be the most portable and still quite efficient, only using around 1–2% of the system CPU resources on average, which is considered reasonable. The range of context signals which are thought to be

| Statistic | Summary |
|---|---|
| processes | number of monitored process including children. |
| cpu (%) | cpu utilisation of the process. The CPU time divided by the time the process has been running (cputime/realtime ratio). |
| mem (%) | ratio of the processes' resident set size to the physical memory on the machine. |
| rss (kB) | resident set size, the non-swapped physical memory that a task has used. |
| size (kB) | approximate quantity of swap space that would be required if the process were to dirty all writeable pages and then be swapped out. |
| sz | size in physical pages of the core image of the process. This includes text, data, and stack space. |
| vsz (KB) | virtual memory size of the process. |
| num_files | total number of files reported by lsof. |
| num_reg | number of regular files. |
| num_dir | number of directories. |
| num_chr | number of character devices. |
| num_ipv4 | number of IPv4 sockets. |
| num_sock | number of sockets of unknown domain. |
| num_unix | number of unix domain sockets. |
| num_unknown | number of unclassified sockets (not reg, dir, ...) |

TABLE 1
Statistics collected by `process_monitor`. Fourteen context signals are collected in total potentially interesting are logged for later analysis and are summarised in Table 1. As well as being feasible to gather, these context signals all relate to a process's interaction with the operating system, and other local or remote processes. Further technical details about the platform set-up are provided in Appendix B.

### 4.2 Normal Usage

The levels of the collected signals need to be examined over a range of different server activities in order to establish which ones are useful for detecting process anomalies. Data is therefore collected over several normal usage scenarios, with the aim of getting the testbed FTP server to behave as if a real FTP server is running on a production network. Ideally, if `strace` and `process_monitor` were installed on a production FTP server then the logs collected by them would be identical to those collected on the testbed FTP server. Realistically, there is a trade-off between the fidelity of the testbed logs and time spent in building and operating the testbed. The methodology employed here allows for largely automated reproduction of normal usage from previously gathered logs of real FTP servers, i.e. public-domain datasets of FTP client-server interactions are used to provide samples of normal usage. The dataset used is a subset of LBNL-FTP-PKT [3], which contains all incoming anonymous FTP connections to public FTP servers at the Lawrence Berkeley National Laboratory over a ten-day period. The dataset is available from the Internet Traffic Archive [3], and its traces, which provide a rich source of normal usage sessions, contain connections between 320 distinct FTP servers and 5832 distinct clients. The traces for one FTP server (IP 131.243.2.12) on the 10/01/03 and 11/01/03 are used in the experiments presented here, providing a total of 340 traces for 76 distinct clients. This particular FTP server is selected as it was running `wuftpd` 2.6.2-1, a similar version to the testbed server used here, and also because it has the second highest number of connections.

In all cases, FTP server activity is produced by the interaction of FTP clients, i.e. an FTP client connects to a server and initiates an FTP session in which it issues FTP requests according to the FTP protocol (defined in RFC959) before disconnecting from the server. Out of the 340 traces used, eight are empty, containing no FTP requests, and many of the other 332 are characterised by USER and PASS commands, followed by an optional STAT, then a series of PORT commands, often lasting tens of minutes, finishing with an optional QUIT. One session of each of these is included in the normal usage dataset and the others (278) are discarded so as not to bias the data.

For the testing of process anomaly detection systems, which is concerned primarily with the type of behaviour and not its frequency, only one example of a typical normal session is necessary. Duplicate sessions are therefore removed from the data, i.e. the original FTP sessions are examined, largely by hand, and sessions which contain the same commands with very similar relative timings (no more than around a second) are removed. These sessions are usually seen coming from the same hosts and appear to be generated by an automated FTP client repeating the same sessions. Indeed, analysis of the two days' worth of traces shows that many sessions are frequently repeated. Discarding duplicates over the two days leaves 55 different normal usage sessions. Although some information in the traces has been anonymised using Bro [17] to remove private information, it is still possible to reproduce realistic normal usage.

### 4.3 Attack Traces

The publically available `autowux` exploit [2] is used to attack `wuftpd`. This exploit levers a format string vulnerability, in this case related to the SITE EXEC FTP command, in order to obtain a remote root shell on the server by default. It has been seen in the wild in manual attacks and automated attacks such as the Ramen worm [19]. An FTP bounce scan attack [1] is also performed; here, an attacker uses an FTP server as an intermediary to perform a network scan and hide the IP address of their machine.

Detecting Anomalous Process Behaviour 15

| Session Name | Nature of Attack |
|---|---|
| success01 | autowux attack |
| | `# ./autowux -t target -v 2` |
| success02 | autowux attack |
| | `# ./autowux -t target -v 2` |
| | `# uname -a` |
| | `# whoami` |
| | `# ls` |
| | `# exit` |
| success03 | autowux attack |
| | `# ./autowux -t target -v 2` |
| | `# cd /` |
| | `# mkdir .boot` |
| | `# cd .boot` |
| | `# ftp host1` |
| | ` anonymous` |
| | ` (no password)` |
| | ` get autowux.tar.gz` |
| | ` quit` |
| | `# tar -xzvf autowux.tar.gz` |
| | `# exit` |
| success04 | nmap bounce scan |
| | `# nmap -v -P0 -b target host` |
| failure01 | autowux attack |
| | `# ./autowux -t target -v 2 -s 1` |

TABLE 2
The five attack sessions collected for `wuftpd` running on an instrumented Redhat 6.2 server. The first three attacks are variations on the autowux attack. The fourth attack is an FTP bounce scan and the final attack is a failed autowux attack

The syscalls and signal levels for several different `autowux` attacks and one nmap FTP bounce attack are recorded. In each case, the commands given on the attacking client machine are summarised in Table 2. The DNS hostname of the FTP server is "target" and "host" denotes the attacking machine. Commands following the first `autowux` command are those given in the remote shell once it has been opened several minutes after the launch of the attack. The session success01 consists of the `autowux` attack without any commands executed in the remote shell, mainly for comparison with the other attacks (although this is fairly unrealistic). Session success02 and session success03 simulate potential actions an attacker might perform in the shell once it has been opened and are of more interest. Session success02 represents a minimal information gathering excerise, while in session success03 the attacker connects back to the attacking machine via FTP and downloads and untars a file, in this case the `autowux` attack itself. Session **success04** is an nmap FTP bounce attack that was successful for unprivileged ports. An unsuccessful `autowux` attack session, **failure01**, is also performed by specifying the insertion of OpenBSD shellcode on the `autowux` command line. This is seen as important since process anomaly detection systems should not alert against failed attacks otherwise they become vulnerable to diversionary noise attacks such as snot [20], which can be used to hide successful attacks. In total these attacks generated around 40,000 syscalls, and 10,000 readings for each signal.

The combined normal and attack traces form a single dataset (called the `wuftpd` dataset) that can be used to evaluate the classification performance of the anomaly detection systems described here. In order to facilitate comparisons with other systems it is publically available [4].

### 4.4 Signal Analysis

The strategy adopted during signal collection as described above is to collect, from the authors' experience, what might be interesting signals. For the purposes of the research presented here interesting signals are those that vary in a *complex* way across normal usage and attack sessions. Clearly, if a particular signal always has certain values for attack sessions and different values for normal sessions, then it would make sense to use this signal as an indicator of misuse and dispense with anomaly detection algorithms. However, no such signals are expected nor have been found. In order to determine which of the collected signals are potentially interesting an analysis of the gathered data is performed.

When examining the signal levels for all 55 normal usage sessions more closely it becomes clear that the memory related signals are closely correlated, as are a number of the file signals. Closer examination of these two groups leads to the elimination of several of the correlated signals. For example, the `size` and `sz` signals always report the same values. Also, several of the signals have the same small number of levels in general regardless of whether the session is an attack or normal usage, and so these signals are also eliminated. The range of values a signal takes over all the normal and attack sessions can be seen from the scatter plots shown in Figure 2. The upper graph shows the observed signal levels for the number of process children, which is considered uninteresting due to its lack of variation over normal and attack sessions. The lower graph in Figure 2 shows the observed levels for the `rss` memory signal, which is considered interesting since there is considerable variation in signal levels between normal and attack sessions, although there is no clear division of levels and crossover exists between signal levels. A similar plot for all the remaining signals shows that there is also considerable variation in the `num_files` and `num_reg` signals. These two signals together with `rss` are used as the context (danger) signals in the experiments that follow.

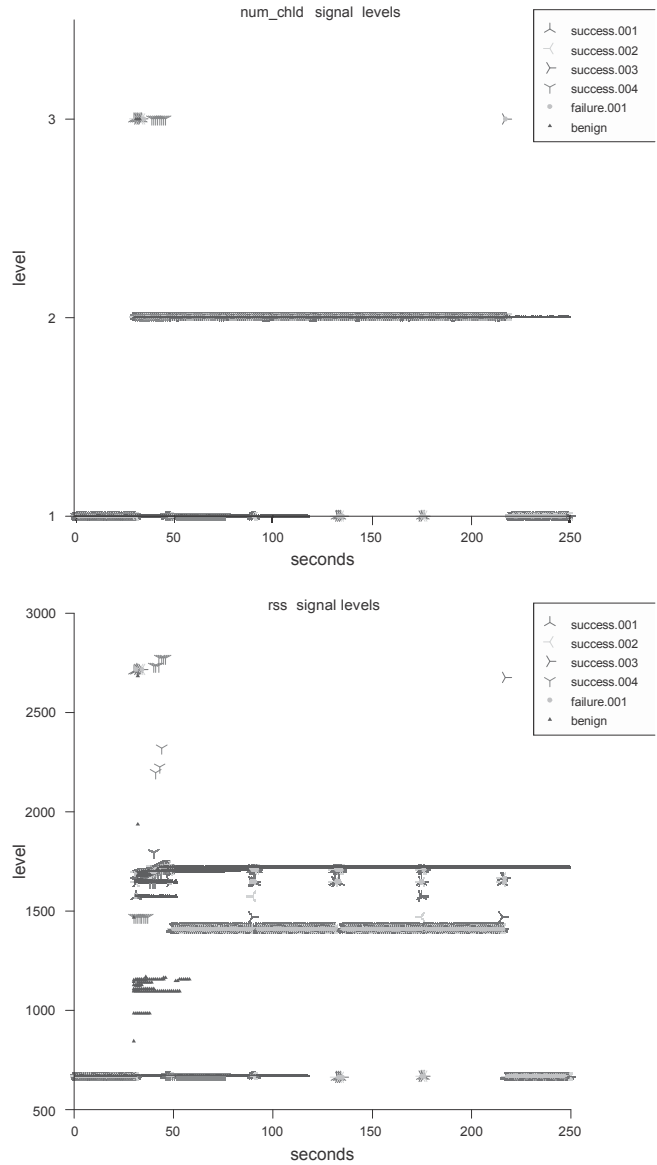

FIGURE 2
Scatter plots of signal levels for all sessions. The top graph shows the levels observed for the number of process children signal. Generally, attack and normal (benign) sessions share similar values for this signal, so it is discarded as uninteresting. The bottom graph shows the levels for the rss memory signal. There is much more variation in this case, with different signal levels being observed for attack and normal sessions, although some values are shared by both classes of session. This signal is considered more useful.

## 5 EXPERIMENTAL PROCEDURES

An important aspect of second generation AISs is their use of multiple sources of input data. This idea has also been advocated by Gao et al. [8] through the notion of gray-boxing. In order to explore these ideas further, three different versions of the `tlr` algorithm are evaluated, named `tlr1`, `tlr2` and `tlr3`. For all three of these algorithms, the syscall number alone is used as antigen. For `tlr1`, the `rss` context signal is used as the danger signal, and immature DCs have a TLR that monitors this signal and is activated by values not seen during training. For `tlr2`, immature DCs have two TLRs, one that responds to the `rss` context signal, and another that is activated by values of the `num_files` context signal not seen during training. If either of these receptors is activated, then the immature DC will develop into a mature DC. For `tlr3`, immature DCs have three TLRs, one for the `rss` context signal, one for the `num_files` context signal, and one for the `num_reg` context signal. In this case, if any of these receptors is activated, the immature DC will change into a mature DC. Hence, the effects of the addition of input data sources can be assessed by comparing the performances of `tlr1`, `tlr2` and `tlr3`.

The use of `process_monitor` and `strace` imposes minimal CPU and memory overheads on the system when gathering data. Timings of syscalls during a single session are preserved and readings of resource usage statistics are taken at regular intervals. One limitation of the dataset is that, when stored as `tcreplay` log files, the overall timings of these sessions relative to each other are not stored. Consequently, when testing AISs with this dataset, an assumption has to be made concerning the relative timings of each session. In the experiments that follow, sessions are assumed to occur sequentially, with a short, uniformly-distributed period of no activity between sessions.

The dynamics of the `tlr` algorithm for a particular FTP session are in part determined by previous FTP sessions. For example, syscalls which are not available during training but which are presented in a normal context during operation, as discussed above, lead to a reduction in the number of TCs specific for such syscalls through peripheral tolerance. Hence, using `tlr` to classify FTP sessions in isolation could produce unrealistic results. A number of *scenarios* are therefore created, consisting of a number of FTP sessions occurring sequentially with a random pause of 1 to 10 seconds between them. This is a simplification of the session timings observed for the actual FTP sessions in the LBNL-FTP-PKT traces, but is necessary in order to reduce the durations of the experiments. (In the LBNL-FTP-PKT traces there are often several minutes between one session and the next, and at other times two sessions overlap.) Here, each scenario takes approximately 50 minutes to run.

Forty scenarios are generated in total, with 20 containing only normal FTP sessions and the remaining 20 containing normal and attack FTP sessions. The latter group should hence be classified as attack sessions. In order to create a scenario the 55 normal sessions are partitoned into two sets of 27 and 28 sessions. One set is used to train `tlr`, and the other set is used to create the scenario on which `tlr` is tested. The two sets are then swapped around, so that the sessions in the testing scenario are used to train `tlr`, and vice versa for the training sessions. In other words, two-fold crossvalidation is performed. The sets are randomly partitioned in this way eight times so that sixteen normal scenarios are generated. The attack sessions are produced in exactly the same way except that an attack session is inserted at a random point in the sequence of normal sessions. The success.001 and success.002 sessions are used six times each, and the success.003 and success.004 session four times each, making a total of 20 attack scenarios. Four additional normal scenarios are also created as just described by inserting the failed attack session at a random point in a sequence of normal sessions. Together with the 16 normal scenarios already generated, this makes a total of 20 normal scenarios.

In order to provide a comparison with the results for the `tlr` algorithm, several other classifiers are also tested with each of the 40 scenarios. First, as a baseline comparison, a classifier that uses a whitelist of acceptable syscalls is used. A syscall policy is generated for a process by recording the syscalls it makes under normal usage, with a permit policy statement entered for all these syscalls. During testing, syscalls are compared to this whitelist and any process that generates unlisted syscalls is classified as under attack. The method is quite realistic considering how current systems such as `systrace` automatically generate a policy.

The whitelist approach is also taken using the signal values `rss`, `num_files` and `num_reg` instead of the syscalls. This generates three simple classifiers, called `sig1`, `sig2` and `sig3`, which extract all the signal levels seen during training. These systems classify a scenario as an attack if any signal levels absent from the whitelist are seen during testing. System `sig1` classifies a scenario as an attack based only on the `rss` signal, system `sig2` uses this and the `num_files` signal, and system `sig3` uses all three signals.

The `tlr` algorithm is also compared to a standard negative selection AIS approach, since for a finite set, a `systrace` whitelist approach and negative selection blacklist approach are logically equivalent and are therefore expected to perform similarly. In order to implement a negative selection algorithm, a slightly altered version of `tlr` (called `tlr-negsel`) is created with the TLRs on immature DCs disabled. In `tlr-negsel` immature DCs are no longer able to respond to signals and hence always differentiate into semimature DCs and never become mature DCs. Consequently, naive TCs are never activated and no alerts are ever produced, since these are generated whenever activated TCs are observed. Therefore, in `tlr-negsel` semimature DCs are forced to express IL-12, effectively turning them into mature DCs, which permits their activation. The absence of semimature DCs in `tlr-negsel` means that no peripheral tolerance of naive TCs by semimature DCs occurs. The only criteria for activation of native TCs is a match between their AgRs and the syscalls being presented by DCs. By disabling peripheral tolerance, negative selection alone (which is used to generate the AgRs) controls the classification performance of `tlr-negsel`.

Finally, the performance of the DCA with the `wuftpd` dataset has also been published in [10], and the results obtained are used here as an additional comparison to the `tlr` algorithm. This is both fair and convenient since both the DCA and the `tlr` algorithm use the `libtissue` framework and make use of the same information in order to process the data. As with the `tlr` algorithm, the DCA uses system call ID numbers to represent antigen and the context signals `rss`, `num_files` and `num_reg` to represent the danger signals.

## 6 RESULTS

The true and false positive rates for each of the classifiers are given in Table 3 below. The true positive rate *TPR* is calculated from:

$$TPR = \frac{TP}{TP + FN}, \quad (1)$$

where *TP* is the number of true positives and *FN* is the number of false negatives. Likewise, the false positive rate *FPR* is calculated from:

$$FPR = \frac{FP}{FP + TN}, \quad (2)$$

where *FP* is the number of false positives and *TN* is the number of true negatives. The table also shows the *g*-mean *G* given by:

$$G = \overline{TPR(1 - FPR)}, \quad (3)$$

| System | TPR | FPR | G |
|---|---|---|---|
| systrace | 0.90 | 0.60 | 0.60 |
| tlr-negsel | 0.90 | 0.60 | 0.60 |
| sig1 | 1.00 | 1.00 | 0.00 |
| sig2 | 1.00 | 1.00 | 0.00 |
| sig3 | 1.00 | 1.00 | 0.00 |
| tlr1 | 0.70 | 0.20 | 0.75 |
| tlr2 | 0.60 | 0.20 | 0.69 |
| tlr3 | 0.75 | 0.15 | 0.80 |
| DCA | 1.00 | 0.83 | 0.41 |

TABLE 3
Classification performance results for the systems implemented. For equal true and false positive costs, `tlr3` is the best performing classifier which provides an overall evaluation of the *TPR* and *FPR*, producing a high value for better classifiers. Note that an equal cost for true and false positives is assumed here. However, in situations where one of these costs is more important than the other, further analysis would be necessary to characterise the relative performance of the classifiers.

Table 3 shows that the `sig1`, `sig2` and `sig3` classifiers perform badly, classifying every scenario as an attack. Consequently, the systems are of no use and this is reflected in the *g*-mean value of 0.00. The `systrace` and negative selection classifiers perform equally, with a high *TPR* of 0.90, but also a high *FPR* of 0.60, meaning that while these classifiers identify 90% of attack scenarios correctly, they also identify 60% of normal scenarios as attacks. The *g*-mean for both of these systems is 0.60. In general, the `tlr` classifiers reduce the *FPR* with an accompanying small reduction in the *TPR*. Compared with the `systrace` and negative selection classifiers, the `tlr1` classifier reduces the *FPR* by 40% to 0.20 with a 20% decrease in the *TPR* to 0.70. The resulting *g*-mean value is 0.75. The `tlr2` classifier has a slightly larger reduction in the *TPR* (30%) to 0.6 but has the same *FPR* as `tlr1`, which gives a *g*-mean value of 0.69. The `tlr3` classifier performs best reducing the *FPR* by 45% to 0.15, while only reducing *TPR* by 15% to 0.75. Its resulting *g*-mean value is 0.80, the highest of all the systems tested. The DCA has a perfect *TPR* of 1.00 i.e. all attack scenarios are correctly classified, but the *FPR* is unacceptably high (0.83), which produces a low *g*-mean value of 0.41.

The `tlr` algorithm, which is unoptimised, uses around 10% of the CPU resources and never more than 8% of the memory resources on the test machine. Generally, CPU usage is only a few percent as cell levels are maintained at a low level during normal usage.

### 6.1 Discussion

The results have provided evidence that the additional context signals `rss`, `num_files` and `num_reg` have the capacity to reduce the false positive rate when used as danger signals in the immune-inspired `tlr` algorithm. In addition, the system performs best when all three of these signals are used in conjunction with the syscall information. However, when the the syscall information is removed (in the case of `sig1`, `sig2`, and `sig3`) and the system is run as a white-list classifier, the IDS is unable to function. This strongly suggests that both types of information source have an important role to play in process anomaly detection.

The results using the negative selection version of the `tlr` algorithm are also poor with regard to the false positve rate. Indeed, Stibor [22] shows that certain matching approaches such as Hamming distance work poorly with negative approaches, introducing an infeasible amount of complexity. Reduction of this complexity by generalisation of the matching criteria results in a significant decrease in the classification performance. Based on these observations, Stibor concludes that negative approaches such as immune-inspired negative selection are unsuitable for real-world anomaly detection problems. Perhaps it would be more correct to say that detection systems based only on negative selection are unsuitable in this context. Used in combination with innate-inspired mechanisms results could be much improved, as demonstrated here.

The `systrace` and negative selection based `tlr-negsel` classifiers give the same classification results in terms of true and false positives. Examination of the results for each scenario shows that both systems also classify exactly the same scenarios as normal and anomalous. This is expected and is a good test for determining the completeness of coverage for the negative selection algorithm. As mentioned previously, the `systrace` whitelist approach and negative selection blacklist approach are in principle equivalent for a finite set of antigen. This is because a blacklist is the complement set of a whitelist, so determining whether an antigen is on the whitelist or not on the blacklist are the same. However, the way negative selection algorithms usually generate a blacklist is stochastic as a dynamically changing set of cells is used. Furthermore, antigen are matched to the blacklist by random encounters between TCs and antigen, so there may be some errors. When implementing the negative selection classifier here, the aim is to reduce these errors as much as possible, otherwise they might influence true and false positive rates. The parameters are hence selected to create a high turnover of naive TCs, which allows them to inspect many semimature and mature DCs and the antigen they present.

The `tlr` algorithm also performs much better than the DCA with regard to the false positive rate. In [10] the chief limitation of the DCA with the `wuftpd` dataset is cited as a requirement for much larger volumes of antigen data, which is connected with the stochastic nature of the DCA's antigen sampling process. Also, the DCA is a much finer-grained system, classifying data on a *per process* basis. This means that, theoretically, the process responsible for the anomalous behaviour can be determined. However, the `wuftpd` dataset does not provide the names of the processes so there is no way in which this information can be utilized to the DCA's advantage.

## 7　CONCLUSIONS

A novel host-based IDS (the `tlr` algorithm) has been presented in this paper. The architecture incorporates mechanisms inspired by both the innate and adaptive biological immune systems, in particular the interactions between two classes of immune cell, DCs and TCs. The performance of the system has been evaluated on a realistic process anomaly detection problem and compared to those of several other classifiers. It was found that the use of the innate immune system mechanisms employed in `tlr` contributed to a decrease in the false positive rate and produced a better overall performance compared to policy-based methods, negative selection approaches, and an alternative DC-based AIS system (the DCA). The paper has also shown how runtime information such as memory and file usage levels can be used in combination with system call information to enhance detection capability. This suggests that such context signals might prove useful in improving the detection capabilities of other IDSs. Indeed, this work has employed novel context signals with a novel algorithm, so it could be the case that more well-established algorithms might perform as well as or better than `tlr` using these context signals. Further research is needed to implement and test such algorithms, although it is unclear how traditional approaches would combine the multiple data sources which is one of the hallmarks of this work. More sophisticated algorithms such as support vector machines or techniques from multisensor data fusion would seem more applicable in this sense.

The generation of the `wuftpd` dataset showed that the collection and usage of runtime statistics as sources of external signals for second generation AISs is possible. Additionally, the use of publically-available datasets to generate the `wuftpd` dataset proved to be an effective methodology. Using part of the LBNL-FTP-PKT data, as well as data from other repositories, this technique could be employed to create a larger database of context signals, which would aid research into the use of these signals as input data for IDSs.

In this work, failed attack sessions are classed as normal, but it may be argued that it is useful and necessary for failed attacks to be registered in order to alert an administrator of suspicious activity. Future work could therefore extend the IDS described here to include a third failed attack class rather than just the normal or abnormal data categories.

Finally, the implementation of the `tlr` and the other second generation AISs more generally showed the feasibility of using `libtissue` to implement AISs as multiagent systems, and of applying them to real-world problems.

## ACKNOWLEDGMENTS

This research is supported by the EPSRC (GR/S47809/01) and HP Labs.

| Parameter Name | Value | Description |
| --- | --- | --- |
| max_antigen | 1000 | maximum number of stored antigen |
| max_cytokines | 3 | maximum number of stored signals |
| max_cells | 10000 | maximum number of cells |
| cell_update_rate ($\mu secs$) | 100000 | rate at which cells are updated |
| antigen_multiplier | 10 | number of copies of each antigen stored |
| num_cells 1 | 100 | number of immature DCs |
| cell_lifespan 1 | 100 | number of iterations an immature DC lives for |
| num_antigen 1 | 100 | maximum number of antigen stored by an immature DC |
| num_antigen_receptors 1 | 10 | number of antigen an immature DC can store |
| num_antigen_producers 1 | 100 | number of antigen an immature DC can produce |
| num_cytokine_receptors 1 | 3 | number of signals an immature DC can respond to |
| antigen_producer_action_time | 10 | number of iterations an immature DC presents antigen for |
| num_cells 2 | 100 | number of naive TCs |
| cell_lifespan 2 | 10 | number of iterations a naive TC lives for |
| num_cell_receptors 2 | 1000 | number of DCs a TC can bind with per iteration |
| num_vr_receptors 2 | 100 | number of antigen a TC can match per iteration |
| cell_lifespan 3 | 100 | number of iterations a semimature DC lives for |
| cell_lifespan 4 | 100 | number of iterations a mature DC lives for |
| cell_lifespan 5 | 100 | number of iterations an activated TC lives for |
| probe_rate ($\mu secs$) | 1000000 | rate at which cell population levels are sampled |

TABLE 4
The libtissue parameter settings used for tlr. For tlr1 and tlr2 the max_cytokines and num_cytokine_receptors_1 parameters are set to 1 and 2 respectively

## A  LIBTISSUE PARAMETER SETTINGS

The libtissue parameter values used for tlr are shown in Table 4.

## B  PLATFORM TECHNICAL DETAILS

A Redhat 6.2 ISO image is downloaded from an official Redhat mirror and used to install a vanilla server as a VMware guest on the testbed. The default wuftpd FTP server package (2.6.0-3) is replaced with a separately downloaded wuftpd 2.6.0-1 RPM. This is necessary as although the wuftpd package in the original Redhat 6.2 distribution has the SITE EXEC vulnerability, Redhat replaced this distribution with a "respin" (read "second edition") which contains a patched `wuftpd` 2.6.0-3. Once installed, the default configuration of `wuftpd` is slightly modified to make it vulnerable.

　In the default installation, `wuftpd` is started at boot time wrapped by the inetd super-server. This makes monitoring more complex as in this configuration the FTP server is only started once a connection has been established through inetd. However, monitoring is technically easier when a process is running continuously. Therefore, the FTP server is disabled in inetd by commenting out the appropriate FTP service entries in /etc/inetd.conf. An init script is written to start the FTP server as a standalone server, running continuously, at boot time. In order to gather the actual data, i.e. process syscall information and context signals, the target system is instrumented: the FTP server executable is wrapped with `strace` [18], which logs all the syscalls made by `wuftpd` and its children.